\documentclass[review]{elsarticle}

\usepackage{lineno, hyperref}
\modulolinenumbers[5]
\journal{Artificial Intelligence Modeling for Dynamical Problems}
\usepackage{tabularx}
\usepackage{textcomp}
\usepackage{graphics}
\usepackage{epsfig}
\usepackage{amssymb,amsmath}
\usepackage{amsfonts}

\usepackage{color}
\usepackage{epstopdf}
\usepackage{multirow}
\usepackage{physics}
\usepackage{float}
\usepackage{mathtools}
\usepackage{amsthm}
\usepackage{caption}
\usepackage{subcaption}
\usepackage{multirow}
\usepackage{booktabs}

\newtheorem{theorem}{Theorem}[section]

\newtheorem{definition}[theorem]{Definition}

\catcode`@=11 \@addtoreset{equation}{section}
\renewcommand\theequation{\thesection.\@arabic\c@equation}
\catcode`@=12

\makeatother

\begin{document}
	
	\begin{frontmatter}
		
		\title{\textbf{Comparative Study of Bending Analysis using Physics-Informed Neural Networks and Numerical Dynamic Deflection in Perforated nanobeam}}

		
		
 
       \author{Ramanath Garai\corref{cor1}}
        \ead{ramanathgarai12@gmail.com}
       \cortext[cor1]{Corresponding author. Email: ramanathgarai12@gmail.com}

		\author[]{Iswari Sahu}
        \ead{sahuiswari920@gmail.com}
			
        \author[]{S. Chakraverty}
	    \ead{sne\_chak@yahoo.com}
        
		\address{Department of Mathematics,
			National Institute of Technology Rourkela, Rourkela-769008,
			{\sc India}.}
		
		%%%%%%%%%%%%%%%%%%%%%%%%%%%%%%%%%%%%%%%%%%%%%%%%%%%%%%%%%%%%%%%%%%%%%%%%%%%%%%%%%%%%%%%%%%%%%%%%%%%%%%%%%%%%%%%%%%%%%%%%%%

\begin{abstract}
In this chapter, we investigate the bending behavior of a perforated nanobeam subjected to sinusoidal loading using an efficient and computationally robust Physics-Informed Functional Link Constrained Framework with Domain Mapping (DFL-TFC) method. Our aim is to determine the relationship between static bending response and dynamic deflection of a perforated nanobeam for various perforation cases. The static bending is obtained using the FL-TFC with Domain mapped method, whereas dynamic deflection is determined using the Galerkin method. The proposed approach employs the theory of functional connections (TFC) to systematically embed governing differential equation constraints into a constrained expression (CE), which exactly satisfies all prescribed initial and boundary conditions (ICs and BCs) and domain of differential equation is mapped to domain of orthogonal polynomials.
Within this framework, the free function appearing in the constrained expression is expressed through a functional link neural network (FLNN). The cost is minimized by the mean square residual of DE, allowing training without requiring complex deep network architectures. Relationship between static and dynamic defection of simply-supported (S-S) perforated nanobeams has been investigated here. FL-TFC with Domain mapped method eliminates the need for deep and complex neural network architectures while ensuring accuracy, efficiency, and strict satisfaction of boundary conditions as compared to standard PINN.
\end{abstract}

%%%%%%%%%%%%%%%%%%%%%%%%%%%%%%%%%%%%%%%%%%%%%%%%%%%%%%%%%%%%%%%%%%%%%%%%%%%%%%%%%%%%%%%%%%%%%%%%%%%%%%%%%%%%%%%%%%%%%%%%%%


\end{frontmatter}

%\linenumbers

%%%%%%%%%%%%%%%%%%%%%%%%%%%%%%%%%%%%%%%%%%%%%%%%%%%%%%%%%%%%%%%%%%%%%%%%%%%%%%%%%%%%%%%%%%%%%%%%%%%%%%%%%%%%%%%%%%%%%

\section{Introduction}
A beam is a common structural element whose length is much larger than its cross-sectional dimensions. Because of their ability to resist bending loads efficiently, beams are widely used in many engineering fields such as civil, mechanical, marine, and aerospace engineering, as well as in micro- and nanoscale structures. In some applications, beams are designed with regularly spaced holes, known as perforated beams. These perforations are often described using a filling ratio that represents the proportion of solid material within a periodic segment. While the inclusion of holes can reduce weight and material usage, it also changes the stiffness and mass distribution of the structure, which can be advantages in some structural applications.

Luschi and Pieri \cite{luschi2012simple,luschi2014analytical} presented an analytical model to study the resonance frequencies of perforated beams. They derived closed-form expressions for the equivalent bending and shear stiffness of beams with regularly spaced square perforations and used them to evaluate the resonance behavior of slender beams. Abdelrahman et al. \cite{abdelrahman2022static} studied the static bending behavior of perforated nanobeams considering microstructure and surface energy effects. Vadivelu et al. \cite{vadivelu2025flexural} examined the flexural behavior of perforated cold-formed steel beams using experimental and numerical methods.

Physics informed neural network was first introduced by Raissi et al. \cite{raissi2019physics}. Here the residual of the governing differential equation is embedded in the loss function during training process. A deep TFC \cite{leake2020deep} framework was developed, which combines neural networks with TFC to solve PDEs ranging from linear to nonlinear cases. Schiassi et al. \cite{schiassi2020extreme} combined Theory of Functional Connection (TFC) with Extreme Learning Machine (ELM) method to solve parametric differential equations. Kumar et al. \cite{10389876} solved Degasperis-Procesi (DP) equation, a challenging nonlinear PDE in computational fluid dynamics for wave propagation, using PINN for boundary value problem.
Li and Bai \cite{li2024physics} present a physics-informed neural network approach for solving non-smooth dynamic problems in friction-induced vibration. A PINN-based approach was employed in \cite{fallah2024physics} to investigate the bending and free vibration behavior of three-dimensional functionally graded porous beams resting on an elastic foundation. Sahoo et al. \cite{sahoo2025physics} implement Physics-Informed Neural Networks (PINNs) to solve large membrane vibration equations, verifying results against ground truth simulations to demonstrate superior performance. Dan et al. \cite{dan2025physics} developed an improved PINN framework for general cable vibration problems by introducing a preferred hardsoft boundary constraint strategy, sine activation function, hierarchical gradient loss with adaptive weighting, and optimized training schemes to accurately capture free vibration responses of cables with bending stiffness. The Dynamic Boundary Condition (DBC) method \cite{martinez2026physics} improves PINN accuracy by dynamically incorporating information from earlier training stages during optimization. It reduces trivial solutions and misleading predictions, demonstrating improved performance on the optical ray equation and Lorenz attractor, with potential applications to PDE problems. 

In the present work, we establish a relationship between the non-dimensional static and dynamic deflections. 
The static deflection is computed using the Functional Link Theory of Functional Connection with Domain mapped method, 
while the dynamic deflection is obtained via Galerkin method. 
For fixed values of filling ratio ($\alpha$), number of rows of holes ($N$), and non-local parameter ($\bar{\alpha}$), the ratio of dynamic to static deflection remains constant across the domain. 
The rest of the chapter is organized as follows: 
Sec. \ref{sec:prelim} presents the preliminaries containing FL-TFC with Domain mapped method and hole-patterned nanobeam model, 
Sec. \ref{sec:derivation} discusses about derivation of governing equation, Sec. \ref{sec:comparison} verifies our proposed method and results, 
Sec. \ref{sec:results} discusses new findings and analyzes the relationship between static and dynamic deflections, 
and Sec. \ref{sec:conclusion} presents the conclusion.

\section{Preliminaries}\label{sec:prelim}

\subsection{Hole-Patterned nanobeam Model}
\begin{figure}[h]
	\centering
	\includegraphics[height=2.7in]{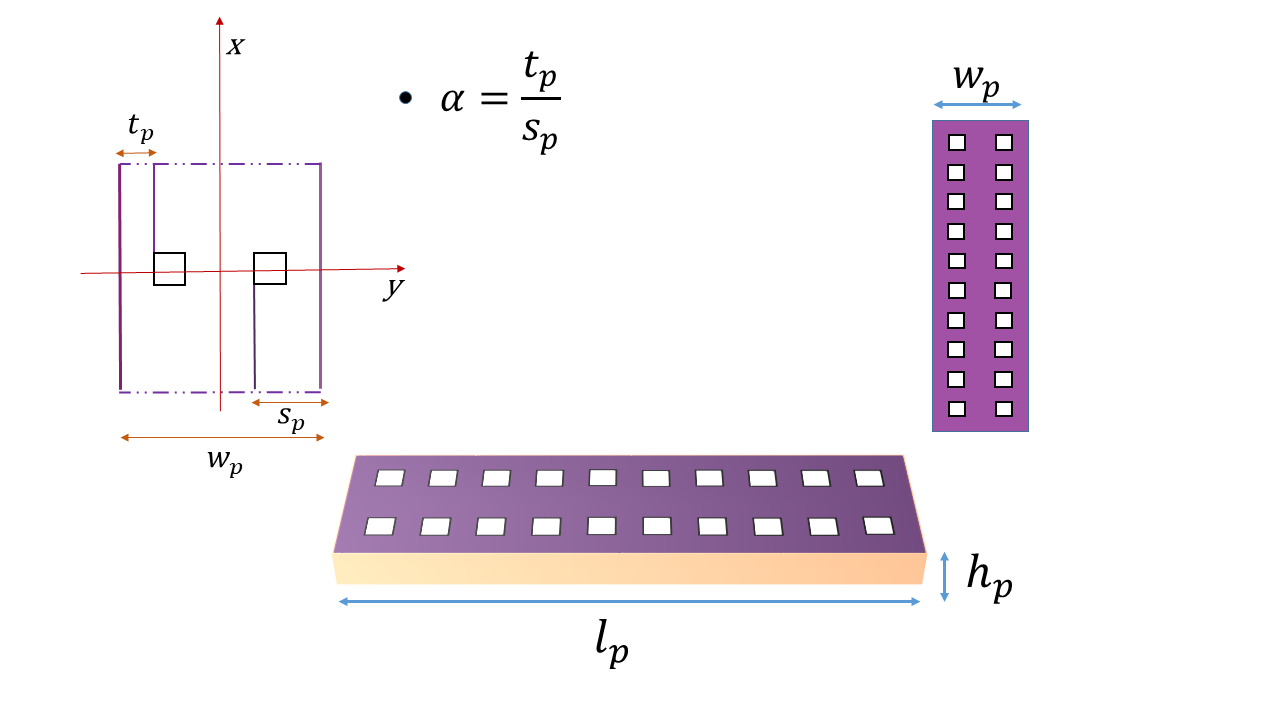}
    \caption{ Perforated nanobeam structural model }
	\label{bookchfig1}
\end{figure}

A perforated nanobeam is a structural member where some part of the material is removed in a regular pattern to change its mechanical behaviour. Because of these perforations, the stiffness, mass distribution and overall response of the nanobeam change. In this study, the nanobeam contains square-shaped perforations arranged periodically along the nanobeam length, forming a grid-like pattern.\\ 
The geometric layout of the perforated nanobeam is presented in Fig.~\ref{bookchfig1}. 
The nanobeam possesses a total length $l_p$, width $w_p$, and thickness $h_p$. 
Along the axial direction of the nanobeam, square perforations are introduced in a periodic manner. The perforation pattern is described by spatial period $s_p$ and period length $t_p$. 
Hence, the side length of each square hole can be written as $s_p - t_p$, while $N$ denotes the number of perforations distributed along the nanobeam length. The ratio $t_p/s_p$ is commonly referred to as the filling ratio, which is expressed as  $\alpha$ \cite{luschi2014analytical,eltaher2018resonance}:
	\begin{equation}
		\alpha = \frac{t_p}{s_p},  \quad 0 \leq \alpha \leq 1 .
	\end{equation}
	  When $\alpha = 1$, the nanobeam corresponds to a completely solid configuration with no perforations. For intermediate values $0 < \alpha < 1$, the nanobeam contains periodically distributed holes, indicating a partially filled structural domain. In the limiting situation $\alpha = 0$, the nanobeam approaches an idealized fully perforated configuration where the material within each periodic cell vanishes. Therefore, the filling ratio can be defined as
	$$
	\alpha =
	\begin{cases}
		0 & \text{Completely Perforated (limiting case)}, \\
		(0,1) & \text{Partially Filled}, \\
		1 & \text{Fully Solid}.
	\end{cases}
	$$

The removal of material from a perforated nanobeam changes the effective bending stiffness as well as mass and rotary inertia of the nanobeam. Therefore, modified expressions are required to represent these quantities for the perforated nanobeam.

Accordingly, the effective bending stiffness of the perforated nanobeam may be written as \cite{luschi2014analytical},
\begin{equation} 
		[EI]_{\text{eq}} = [E I]
		\left[ \frac{\alpha (N+1)(N^2 + 2N + \alpha^2)}
		{\left(1 - \alpha^2 + \alpha^3\right)N^3 + 3\alpha N^2 + \left(3 + 2\alpha - 3\alpha^2 + \alpha^3\right)\alpha^2N + \alpha^3} \right],
	\end{equation}
where $E$ is the Young's modulus, $I$ is the moment of inertia of the nanobeam. 

In addition, the presence of perforations also affects the mass characteristics of the nanobeam. As a result, the effective mass per unit length of a perforated nanobeam differs from that of a completely solid nanobeam. Similarly, the moment of inertia per unit length is also modified due to removal of material. These quantities can therefore be expressed in their modified forms as \cite{luschi2014analytical},
\begin{equation}
	[\rho A]_{eq} = [\rho A]   \left[ \frac{[1 - N(\alpha - 2)]\alpha}{N + \alpha} \right],
\end{equation}
\begin{equation}
	[\rho I]_{\text{eq}}= [\rho I]  \left[ \frac{\alpha \left[ (2 - \alpha)N^3 + 3N^2 - 2(\alpha - 3)(\alpha^2 - \alpha + 1)N + \alpha^2 + 1 \right]}{(N + \alpha)^3} \right].
\end{equation}
Here, $\rho$ represents the density of the nanobeam, while $A$ denotes its cross-sectional area.

\section{Derivation of the Governing Equation}\label{sec:derivation}

\subsection{Static Deflection Case }
In the Euler-Bernoulli beam theory, deflection of the beam at any point along its length $l_p$ is represented by $w_p(x)$. For small deformations, displacement components in the $X$, $Y$, and $Z$ directions are expressed as \cite{reddy2007nonlocal,fernandez2017vibrations}: 
\begin{equation}
    P_1(x,z,t) = v_p(x,t) - z \frac{\partial w_p(x,t)}{\partial x},
\end{equation}
\begin{equation}
    P_2(x,z,t) = 0,
\end{equation}
\begin{equation}
    P_3(x,z,t) =  w_p(x,t).
\end{equation}
Here, $v_p$ and $w_p$ denote the axial and transverse deflections of the beam, respectively. Because axial displacement along the neutral axis is much smaller than transverse deflection and associated rotation, it is considered negligible and therefore omitted from the analysis. 
The nonzero strain can be
described as
\begin{equation}
    \epsilon_{xx} = \frac{\partial P_1(x,z,t)}{\partial x}.
\end{equation}

The quantities $\sigma_{xx}$, $N_r$, and $M_r$ represent axial stress, resultant normal force, and bending moment, respectively, which are defined as follows: 
	\begin{subequations} 
		\begin{equation}
			\sigma_{xx} = E \epsilon_{xx}  =  E \left( \frac{\partial v_p(x,t)}{\partial x} -z \frac{\partial^2 w_p(x,t)}{\partial x^2} \right),  \label{p2eq14a}   
		\end{equation}
		\begin{equation}
			N_r = \int_A \sigma_{xx}dA = \int_A E \epsilon_{xx}dA ,
		\end{equation}
		\begin{equation}
			M_r = \int_A z \sigma_{xx} \, dA = \int_A E z \epsilon_{xx} \, dA .
		\end{equation}
	\end{subequations}
    
According to Eringen's nonlocal theory, the bending moment can be written as \cite{Eringen1983,reddy2007nonlocal}:
\begin{equation} \label{bookeq3.2}
    M_r - (e_0 a)^2 \frac{\partial^2 M_r}{\partial x^2} = -[EI]_{eq} \frac{\partial^2 w_p(x)}{\partial x^2}.
\end{equation}
The nonlocal parameter is written as $(e_0 a)$, where $e_0$ is a material-dependent constant and $a$ denotes the internal characteristic length of the material.
The equation of motion from the Lagrange equation is written as:
\begin{equation}  \label{bookeq3.3}
\frac{\partial ^2 M_r}{\partial x^2} = - q(x).
\end{equation}

Differentiating twice Eq. (\ref{bookeq3.2}) and then using Eq. (\ref{bookeq3.3}), we get the final governing differential equation:
\begin{equation} \label{bookeq3.4}
    [EI]_{eq} \frac{\partial^4 w_p(x)}{\partial  x^4} = q(x)  - (e_0 a)^2  \frac{\partial ^2 q(x)}{\partial  x^2}.
\end{equation}

On modifying Eq.~(\ref{bookeq3.4}), the governing equation of motion for bending of a perforated nanobeam under a sinusoidal load is obtained. The nonlocal effect is incorporated in the formulation, and the resulting equation can be written as \cite{kafkas2025size}:
\begin{equation} \label{bookeq3.9}
    [EI]_{eq} \frac{d^4 w_p(x)}{d  x^4} = q(x)  - (e_0 a)^2  \frac{d ^2 q(x)}{d  x^2}.
\end{equation}

To generalize the new findings and reduce the number of parameters, a set of parameters is introduced as follows:
$$ X=\frac{x}{l_p} , \ \ W_p(X) = \frac{w_p(x)}{l_p} , \ \ Q(X) = \frac{q(x)}{l_p}.$$

After applying the above non-dimensional terms, Eq. (\ref{bookeq3.9}) is obtained as:
\begin{equation} \label{bookeq3.5}
    [EI]_{eq} \frac{1}{l_p^3} \frac{d^4 W_p(X)}{d X^4} = l_p Q(X) - (e_0 a)^2\frac{1}{l_p} \frac{d^2 Q(X)}{d X^2}.
\end{equation}

Applying the external sinusoidal load $Q(X) = q_0 sin(\pi X)$, the equation (\ref{bookeq3.5}) is written as: 
\begin{equation} \label{bookeq3.6}
     \frac{d^4 W_p(X)}{d X^4} = \frac{l_p^4 q_0}{EI} \left[ \frac{sin(\pi X)}{P_1}+ \frac{\bar \alpha^2 \pi^2 sin(\pi X) }{P_1} \right].
\end{equation}

Further, the above equation is expressed as:
\begin{equation} \label{bookeq3.7}
     \frac{d^4 \overline W_p(X)}{d X^4} =  \left[ \frac{sin(\pi X)}{P_1}+ \frac{\bar \alpha^2 \pi^2 sin(\pi X) }{P_1} \right], 
\end{equation}
where $\overline W_p(X) = \frac{E IW(X)  }{l_p^4 q_0}, \bar \alpha^2 = (\frac{e_0a}{l_p})^2, \ \text{and} \ P_1 =	\left[ \frac{\alpha (N+1)(N^2 + 2N + \alpha^2)}
		{\left(1 - \alpha^2 + \alpha^3\right)N^3 + 3\alpha N^2 + \left(3 + 2\alpha - 3\alpha^2 + \alpha^3\right)\alpha^2N + \alpha^3} \right].$
        
When the nanobeam is simply supported (S-S) and subjected to a transverse load $Q(X)$, the support conditions require zero deflection and zero bending moment at both ends. Thus, at $X = 0$ and $X = 1$, the boundary conditions can be written as:
\begin{equation}
    \overline W_p(X) = 0, \frac{d^2 \overline W_p(X) }{d X^2} = 0.
     \label{bookchBC}
\end{equation}

\subsection{Dynamic Deflection Case }
The governing differential equation describing the dynamic behavior of the perforated nanobeam subjected to an external sinusoidal load is expressed as follows \cite{eltaher2018resonance,garai2026effect}:
\begin{equation}
    \frac{d^2}{dX^2} \left[ P_1  \frac{d^2 \overline W_p }{d X^2} \right] = \bar \alpha^2 \lambda^2 \left[ -P_2  \overline W_p  + P_3 \frac{h_p^2}{12 \times l_p^2 }  \frac{d^2 \overline W_p }{d X^2}    \right] + \lambda^2 \left[ P_2  \overline W_p  - P_3 \frac{h_p^2}{12 \times l_p^2 }  \frac{d^2 \overline W_p }{d X^2}    \right],
    \label{bookcheq3.14}
\end{equation}
where $\lambda^2 =  {\frac{\omega^2 l^4_p \rho A}{E I}} \ \text{is the frequency parameter}, \bar \alpha^2 = (\frac{e_0a}{l_p})^2  \ \text{is non-local parameter}, $ \\
$ P_1 = \left[ \frac{\alpha (N+1)(N^2 + 2N + \alpha^2)}
		{\left(1 - \alpha^2 + \alpha^3\right)N^3 + 3\alpha N^2 + \left(3 + 2\alpha - 3\alpha^2 + \alpha^3\right)\alpha^2N + \alpha^3} \right] $, \\
        $P_2 = \left[ \frac{[1 - N(\alpha - 2)]\alpha}{N + \alpha} \right]  $ and \\
        $P_3 = \left[ \frac{\alpha \left[ (2 - \alpha)N^3 + 3N^2 - 2(\alpha - 3)(\alpha^2 - \alpha + 1)N + \alpha^2 + 1 \right]}{(N + \alpha)^3} \right].$

The above Eq. (\ref{bookcheq3.14})  represents the relationship between the material properties, geometric characteristics, and the resulting dynamic deformation of the nanobeam.
This Eq.~(\ref{bookcheq3.14}) is solved together with the above-mentioned boundary conditions (Eq.~(\ref{bookchBC})) using Galerkin method. From this procedure, a non-dimensional frequency parameter $\lambda$ is obtained. After determining the frequency parameter $\lambda$, the corresponding first mode shape, represented by the dynamic deflection $\overline W_p$, is obtained.
Then, for the simplification and new findings, it is multiplied by 100, and it is expressed as:
$$\widetilde{W_p}^d(X) = 100 \times \overline W_p(X), \  \ X \in [0,1].$$

\section {Physics-Informed Functional Link Constrained Framework with Domain Mapping (DFL-TFC) Method } \label{sec-FLTFC}
 The governing differential equation \ref{bookeq3.7} is 
\begin{equation} \label{bookchEq4.1}
     \frac{d^4 \overline W_p(X)}{d X^4} =  \left[ \frac{sin(\pi X)}{P_1}+ \frac{\bar \alpha^2 \pi^2 sin(\pi X) }{P_1} \right],
\end{equation}

 with boundary conditions 
\begin{equation*}
    \overline W_p(0) = 0, \frac{d^2 \overline W_p(0) }{d X^2} = 0,
    \end{equation*}

    \begin{equation*}
     \overline W_p(1) = 0, \frac{d^2 \overline W_p(1) }{d X^2} = 0.
     \label{bookchBC1}
\end{equation*}

 In FL-TFC with Domain mapped \cite{sahu2026physics1}, domain of the DE $[0, 1]$ is transformed to standard interval of chebyshev polynomials $[-1, 1]$ \cite{mortari2019high} and expands input features $X$ into a higher-dimensional space using these chebyshev polynomials of order $14$, enabling the model to capture nonlinear relationships more effectively.  The expanded features are then used to construct the neural network approximation. A set of $100$ points within the domain is used for training.  Here, we have implemented functional link neural network (FLNN) \cite{mall2017single} as free function $h(X)$. The CE, derived from the TFC \cite{mortari2017theory, mortari2019multivariate, leake2020multivariate}, combining free functions, switching functions, and support functions (Detailed given in Subsec. \ref{label2.2}), takes the form
 \begin{equation}\label{eq:constrained_expression}
   \widehat{\overline{W}_p}(X) = h(X) + (X-1) h(0) - X h(1) + \frac{2X - 3X^2 + X^3}{6} h''(0) + \frac{X - X^3}{6} h''(1)
 \end{equation}
  The resulting constrained solution (Eq. (\ref{eq:constrained_expression})) is substituted into the governing differential equation \ref{bookchEq4.1} to compute the residual. The residual can be defined as 
\begin{equation}\label{eq:residual}
\mathcal{R}(X) = \frac{d^4 \widehat{ \overline W_p}(X)}{d X^4} -  \left[ \frac{sin(\pi X)}{P_1}+ \frac{\bar \alpha^2 \pi^2 sin(\pi X) }{P_1} \right].
\end{equation}
 Gradients can be computed using automatic differentiation (AD) \cite{paszke2017automatic}. Finally, the neural network parameters are optimized by minimizing the loss using gradient-based optimization algorithms such as L-BFGS. The optimization is performed in five steps, with a maximum of 50 iterations in each step, allowing the model to iteratively learn an approximate solution \( (\widehat{\overline{W}_p}(X))\) that satisfies both the governing equation and prescribed constraints. For simplification, it is multiplied by 100, and it is represented as:
$$\widetilde{W_p}^s(X) = 100 \times \widehat{\overline{W}_p}(X), \  \ X \in [0,1].$$
% Graphical overview of FL-TFC framework is shown in Fig. \ref{fig: PIFLNN}.

% \begin{figure}[H]
%     \centering
%     \includegraphics[width=0.993\linewidth]{PIFLNN diagram.pdf}
%     \caption{Graphical overview of FL-TFC method}
%     \label{fig: PIFLNN}
% \end{figure}

\subsection{How to make CE?}\label{label2.2}

\begin{definition} \label{def. constraint_operator} \cite{leake2020multivariate, sahu2026physics}
An operator, denoted by ${}^{x_k}\mathfrak{C}^i[\cdot]$, is referred to as the \textbf{constraint operator} associated with the $i^{\text{th}}$ constraint of the variable $x_k$. This operator acts on a function $f$ by evaluating it according to the specified constraint condition. For instance, for pointwise value constraints it is defined as
\[
{}^{x_k}\mathfrak{C}^i[f] = f(x_k^i),
\]
where $x_k^i$ represents the location corresponding to the $i^{\text{th}}$ constraint of the variable $x_k$.
\end{definition}

\begin{definition} \label{def. projection_functional} \cite{leake2020multivariate,  sahu2026physics}
The \textbf{projection functional}, denoted by ${}^{x_k}\rho_j$ or $\rho_j(x_k, h(X))$, associated with the $j^{\text{th}}$ constraint of the variable $x_k$ and the free function $h(X)$, is defined as
\[
\rho_j(x_k, h(X)) = k_j - {}^{x_k}\mathfrak{C}^j[h],
\]
where $k_j \in \mathbb{R}$ represents the prescribed value of the $j^{\text{th}}$ constraint.
\end{definition}

Let \(X = (x_1, x_2, \dots, x_n)^\top \in \mathbb{R}^n\) represent an $n$-dimensional input vector. Let us consider a function $f(X)$ subject to $k$ constraints across its domain. $h(X)$ denotes an arbitrary free function. 
In the multivariate theory of functional connections (TFC) framework, 
a constrained expression \(\hat{f}(X, h(X))\) can be constructed 
to automatically satisfy all specified constraints, with the CE provided below  \cite{leake2020multivariate, sahu2026physics, sahu2026physics1}:
\begin{equation}\label{eq. CE formula}
    \hat{f}(X, h(X)) = h(X) +  \mathcal{M} \left( \boldsymbol{\rho}(X, h(X)) \right)\, \Psi_{x_1}\, \Psi_{x_2} \cdots \Psi_{x_n},
\end{equation}
where the components are defined as follows:

\begin{itemize}
    \item \(\mathcal{M}\) is an \(n\)-dimensional tensor, whose elements are computed from the the constraints themselves and linked projection functionals.

    \item If keeping only one index free (e.g., in the \(k^{\text{th}}\) dimension) while setting all others indices to 1, the corresponding tensor elements are given by
    \begin{equation}\label{eq. m_11}
        \mathcal{M}_{1\cdots x_k \cdots 1} = 
    \left\{0, {}^{x_k}\rho_{1},\cdots,{}^{x_k}\rho_{l_{x_k}}\right\}.
    \end{equation}

\item For tensor elements where more than one index is different from 1 (i.e., corresponding to constraints in multiple variables), are obtained as the geometric intersections of the corresponding projection functionals, multiplied by an appropriate sign $(+\,\text{or}\,-)$. Mathematically it is written as,
    \begin{equation}\label{eq. m_ij}
        \mathcal{M}_{i_{x_1}i_{x_2}...i_{x_n}} = (-1)^{n+1}\,{}^{x_1}\mathfrak{C}^{i_{x_1}-1}\bigg[{}^{x_2}\mathfrak{C}^{i_{x_2}-1}\Big[\cdots\left[{}^{x_n}\rho_{i_{x_n}-1}\right]\cdots\Big]\bigg].
    \end{equation}

 \item For each variable $x_k$, the switching function vector corresponding to $l_{x_k}$ constraints is given by
 \begin{equation}
\boldsymbol{\Psi}_{x_k} = \begin{bmatrix}
1 \\ 
^{x_k}\psi_1 \\ 
\vdots \\ 
^{x_k}\psi_{l_{x_k}}
\end{bmatrix} \in \mathbb{R}^{l_{x_k} + 1}.
\end{equation}

\item Each switching function ${}^{x_k}\psi_j$ is constructed as a linear combination of $l_{x_k}$ support functions:
 \begin{equation}\label{eq:switching_fn}
{}^{x_k}\psi_j(x_k) = \sum_{i=1}^{l_{x_k}} \alpha_{ij} \, \mathtt{s}_i(x_k),
\end{equation}
where $\mathtt{s}_i(x_k) = x_k^{i-1}$ represents the monomial support functions.

    \item The monomial support functions $\{\mathtt{s}_i(x_k)\}$ are chosen to be linearly independent over the $x_k$ domain, so that the coefficient matrix remains invertible and is given by
\begin{equation}\label{eq:coeff_matrix}
[\alpha_{ij}]_{l_{x_k} \times l_{x_k}} = \left[ {}^{x_k}\mathfrak{C}^i [\mathtt{s}_j(x_k)] \right]^{-1}.
\end{equation}
The coefficients $\alpha_{ij}$ are determined by imposing the condition that each switching function $\psi_j(x)$ satisfies
\begin{equation}
{}^{x}\mathfrak{C}^i[\psi_j(x)] = \delta_{ij}, \quad \text{for } i,j = 1,2,\dots,k,
\end{equation}

where $\delta_{ij}$ denotes the Kronecker delta. This guarantees that $\psi_j(x)$ is active only at the $j^{\text{th}}$ constraint while remaining zero at all others constraint.
\end{itemize}

\section{Validation of Results}\label{sec:comparison}
To validate the effectiveness of the proposed approach in capturing bending behavior of a nanobeam under external loading conditions, a simply supported perforated nanobeam subjected to a sinusoidal transverse load is analyzed. Since the formulation is expressed in a non-dimensional form, the obtained results do not depend on any particular set of material properties. Therefore, no specific nanobeam material is assumed in the present analysis.

For validation purposes of the static bending deflection of a simply supported perforated nanobeam, the mean square residual loss of FL-TFC with Domain mapped method is derived. It achieves a residual loss of $10^{-11}$. This significant reduction in loss value indicates that the proposed FL-TFC with Domain mapped framework converges efficiently for static bending behavior. In addition, dynamic deflection is evaluated using the Galerkin method. The convergence behaviour of dynamic deflection is presented in Table \ref{bookchVT}. From Table \ref{bookchVT}, it can be observed that dynamic deflection values converge when $n \ge 10$. This confirms the accuracy of the method and validates newly obtained results. $n = 14$ is chosen for further analysis to obtain new results. 

The following set of parameter values is considered for the validation purpose:

S-S BC: $\alpha =0.5,$   $N=2 $, \text{and} $ \bar \alpha = 0.2.$

\begin{table}[H]
	\centering
	\caption{Convergence table of dynamic deflection for the S-S perforated nanobeam with parameters $ N = 2 $, $\alpha = 0.5 $, and $\bar \alpha = 0.2 $ }
	\begin{tabular} {c c c c }
		\hline
		$n$ & $\widetilde{W_p}^d(0.3)$ & $\widetilde{W_p}^d(0.5)$ & $\widetilde{W_p}^d(0.9)$ \\
		\hline
		8	& 108.3092 & 133.8678 & 41.3674  \\
		\hline
		9	& 108.3056 & 133.8730 & 41.3690 \\
		\hline
		10	& 108.3056 & 133.8730 & 41.3690  \\
		\hline
		11	& 108.3056 & 133.8730 & 41.3690 \\
		\hline
		12	& 108.3056 &133.8730 & 41.3690  \\
		\hline
		13	& 108.3056 &133.8730 & 41.3690  \\
		\hline
		14	& 108.3056 &133.8730 & 41.3690 \\
		\hline
        15	& 108.3056 &133.8730& 41.3690 \\
		\hline
		
	\end{tabular} 
    \label{bookchVT}
\end{table}

\section{Results and Analysis}\label{sec:results}
\subsection{Significant Findings}
The bending deflection of a perforated nanobeam under a sinusoidal load is analyzed by the influence of perforation parameters, such as filling ratio and number of holes, and by the non-local parameter. Table \ref{bookchT1} includes non-dimensional static and dynamic deflections with varying filling ratios and numbers of holes. The relationship between non-dimensional static and dynamic deflections is established in this chapter. From Table \ref{bookchT1}, it can easily be seen that for fixed $\alpha$ and $N$, the ratio of dynamic and static deflection is a fixed constant for all $X \in (0,1)$. This means here $\alpha=0.3, N =1$ the ratio constant is $90.6711$ for all $X \in (0,1)$, when $\alpha=0.3, N=2$ the ratio constant is $68.5328$ in $X \in (0,1)$ and so on as given in Table \ref{bookchT1}. It is interesting to note that with this ratio constant, we can say that these two deflections are proportional, and by this ratio constant, one can find the other deflection using another, and the shape or deflection pattern is similar for both, which means both follow a similar pattern curve.  

Also, from Table \ref{bookchT1}, it can be observed that when the filling ratio increases, deflection for both static and dynamic cases decreases. This happens because a higher filling ratio means more material is present in the perforated nanobeam, which increases effective stiffness and therefore reduces deflection. In contrast, static and dynamic deflections increase when the number of holes increases.

\begin{table}[H] 
\centering
\caption{Relationship between the static and dynamic deflection of a simply supported (S-S) perforated nanobeam for different values of $\alpha$ and $N$ with the parameter $\bar{\alpha}=0.2$ } 
\renewcommand{\arraystretch}{1.2}
 \setlength{\tabcolsep}{9pt}
\begin{tabular}{|c|c|c|c|c|c|}
\hline
Filling Ratio $(\alpha)$ & N & Deflection $(\widetilde{W_p})$& Static $(\widetilde{W_p}^s)$ & Dynamic $(\widetilde{W_p}^d)$ & Ratio $=\frac{\widetilde{W_p}^d}{\widetilde{W_p}^s}$\\
\hline

  \multirow{6}{*}{0.3} & \multirow{3}{*}{$1$} & $\widetilde{W_p}(0.3)$ & 1.3534 & 122.7182 &90.6711  \\ 
 &  & $\widetilde{W_p}(0.5)$ & 1.6729 & 151.6880 &90.6711  \\ 
 &  & $\widetilde{W_p}(0.9)$ & 0.5170 &46.8742 &90.6711  \\ \cline{2-6}

 & \multirow{3}{*}{$2$} & $\widetilde{W_p}(0.3)$ & 1.8658 & 127.8701 & 68.5328  \\ 
 &  & $\widetilde{W_p}(0.5)$ & 2.3063 & 158.0562 & 68.5328 \\ 
 &  & $\widetilde{W_p}(0.9)$ & 0.7127 & 48.8420 & 68.5328 \\ \hline

\multirow{6}{*}{$0.5$} & \multirow{3}{*}{$1$} & $\widetilde{W_p}(0.3)$ & 1.1918 & 106.1157 & 89.0357 \\ 
 &  & $\widetilde{W_p}(0.5)$ & 1.4732 & 131.1662 & 89.0357 \\ 
 &  & $\widetilde{W_p}(0.9)$ & 0.4552 & 40.5326 &89.0357  \\ \cline{2-6}

 & \multirow{3}{*}{$2$} & $\widetilde{W_p}(0.3)$ &  1.3866 & 108.3056 & 78.1092 \\ 
 &  & $\widetilde{W_p}(0.5)$ & 1.7139 & 133.8730 &  78.1092\\ 
 &  & $\widetilde{W_p}(0.9)$ & 0.5296 & 41.3690 & 78.1092 \\ \hline

\multirow{6}{*}{$0.7$} & \multirow{3}{*}{$1$} & $\widetilde{W_p}(0.3)$ & 1.1617 & 99.5418 &85.6876  \\ 
 &  & $\widetilde{W_p}(0.5)$ & 1.4359 & 123.0404 &85.6876  \\ 
 &  & $\widetilde{W_p}(0.9)$ & 0.4437 & 38.0216 & 85.6876 \\ \cline{2-6}

 & \multirow{3}{*}{$2$} & $\widetilde{W_p}(0.3)$ & 1.2199 & 100.2720 & 82.2000 \\ 
 &  & $\widetilde{W_p}(0.5)$ & 1.5078 & 123.9430 & 82.2000 \\ 
 &  & $\widetilde{W_p}(0.9)$ & 0.4659 & 38.3005 &  82.2000\\ \hline

\end{tabular}
\label{bookchT1}
\end{table}

 In Table \ref{bookchT2}, the non-dimensional static and dynamic deflections are presented while varying non-local parameter ($\bar \alpha$) and the number of holes ($N$). Here, again, it can be seen that the ratio of the non-dimensional dynamic and static deflections is a fixed constant for fixed $\bar \alpha$ and $N$. For example, when $\bar \alpha =0.1,  N=1$, the ratio constant is $127.3527$, when $\bar \alpha =0.2,  N=2$, the ratio constant is $78.1092$ and so on as listed in Table \ref{bookchT2}. Because this ratio remains fixed, deflection in one case can be determined from the other through simple scaling. In addition, both responses exhibit the same deflection profile, indicating that static and dynamic curves have an identical shape. From Table \ref{bookchT2} one can see that, as the non-local parameter increases, static deflection becomes larger, while dynamic deflection tends to decrease. In contrast, an increasing number of holes causes the non-dimensional static deflection to increase, since removing more material makes the nanobeam more flexible and allows greater bending.

\begin{table}[H]
\centering
\caption{Relationship between the static and dynamic deflection of a simply supported (S-S) perforated nanobeam for different values of $\bar \alpha$ and $N$ with the parameter ${\alpha}=0.5$} 

\renewcommand{\arraystretch}{1.2}
 \setlength{\tabcolsep}{9pt}
\begin{tabular}{|c|c|c|c|c|c|}
\hline
Non-local Parameter ($\bar {\alpha}$) & N & Deflection $(\widetilde{W_p})$ & Static $(\widetilde{W_p}^s)$ & Dynamic $(\widetilde{W_p}^d)$& Ratio $=\frac{\widetilde{W_p}^d}{\widetilde{W_p}^s}$\\
\hline

\multirow{3}{*}{$0.1$} & \multirow{3}{*}{$1$} & $\widetilde{W_p}(0.1)$ &  0.3586& 45.6687 & 127.3527 \\ 
 &  & $\widetilde{W_p}(0.6)$ & 1.1037 & 140.5539 & 127.3527 \\ 
 &  & $\widetilde{W_p}(0.8)$ & 0.6821 & 86.8671  & 127.3527 \\ \hline

\multirow{3}{*}{$0.2$} & \multirow{3}{*}{$2$} & $\widetilde{W_p}(0.1)$ & 0.5296 & 41.3690 & 78.1092 \\ 
 &  & $\widetilde{W_p}(0.6)$ & 1.6300 & 127.3208 &78.1092  \\ 
 &  & $\widetilde{W_p}(0.8)$ & 1.0074 & 78.6886 &  78.1092\\ \hline

\multirow{3}{*}{$0.3$} & \multirow{3}{*}{$3$} & $\widetilde{W_p}(0.1)$ & 0.7813 & 35.8767 & 45.9186 \\ 
 &  & $\widetilde{W_p}(0.6)$ & 2.4047 & 110.4171 & 45.9186 \\ 
 &  & $\widetilde{W_p}(0.8)$ & 1.4862 & 68.2415 & 45.9186 \\ \hline

\multirow{3}{*}{$0.4$} & \multirow{3}{*}{$4$} & $\widetilde{W_p}(0.1)$ &  1.1269& 30.8541 & 27.3792 \\ 
 &  & $\widetilde{W_p}(0.6)$ &  3.4683 & 94.9590 & 27.3792 \\ 
 &  & $\widetilde{W_p}(0.8)$ & 2.1435 & 58.6879 & 27.3792 \\ \hline

\end{tabular}
\label{bookchT2}
\end{table}

\subsection{Parameter-Based Analysis}
\begin{figure}[H]
  \centering
  \begin{subfigure}[b]{0.49\linewidth}
  \includegraphics[width=\linewidth]{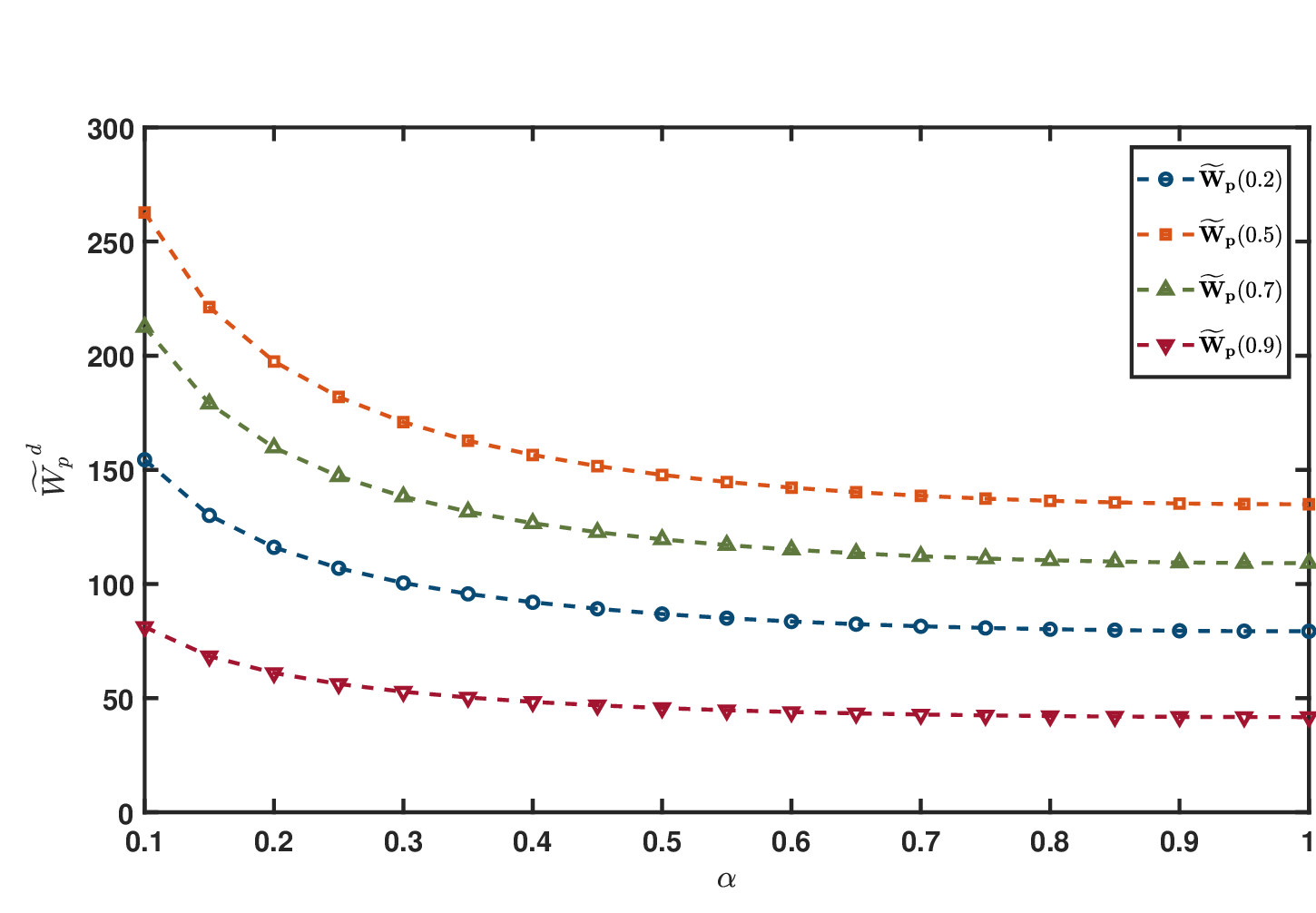}
\caption{ Variation of dynamic deflection along the filling ratio $\alpha$ with the parameters $N=1$, $\bar \alpha = 0.1$}
  \end{subfigure}
  \begin{subfigure}[b]{0.49\linewidth}
\includegraphics[width=\linewidth]{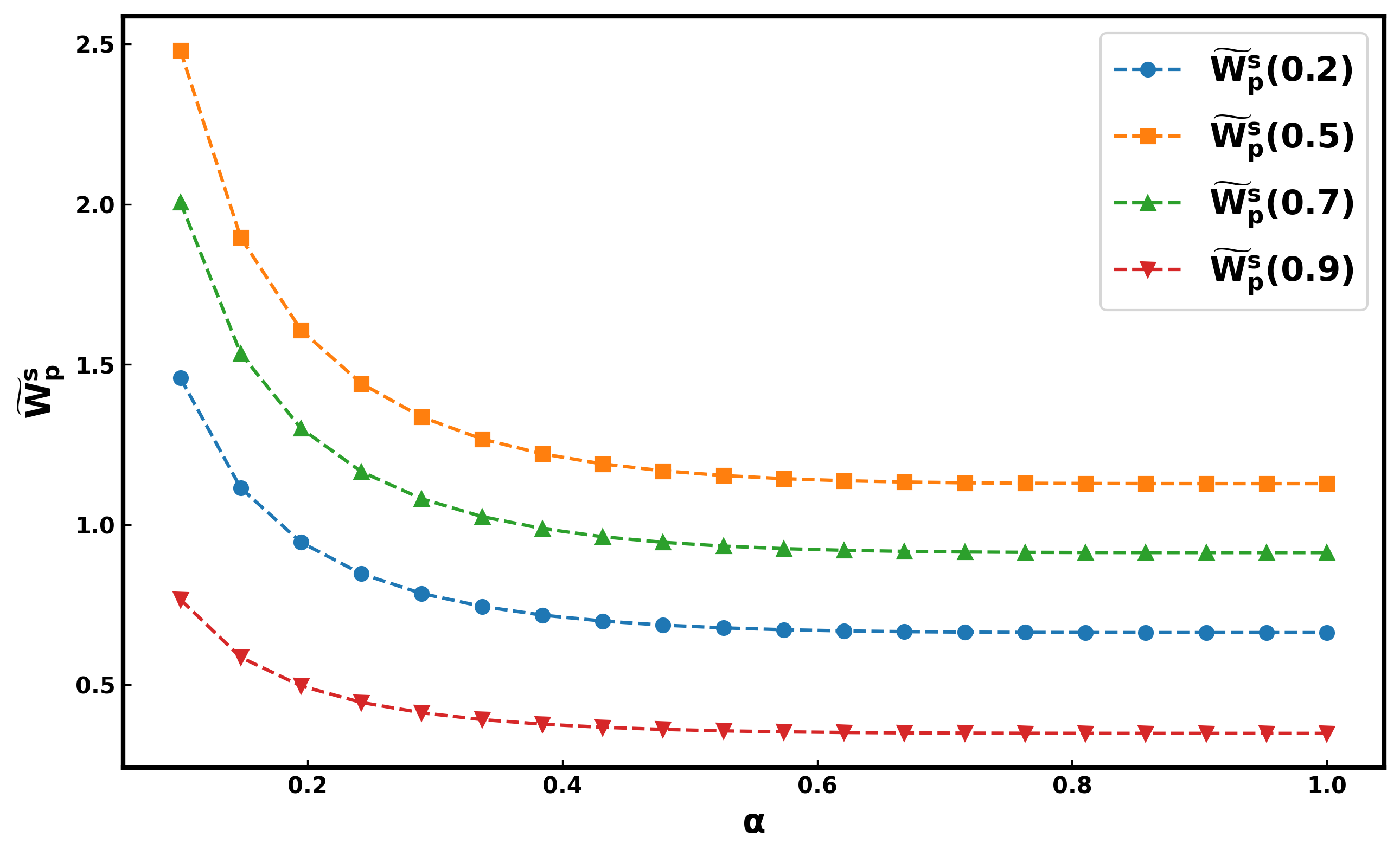}
\caption{Variation of static deflection along the filling ratio $\alpha$ with the parameters $N=1$, $\bar \alpha = 0.1$}
  \end{subfigure}
  \caption{Static and dynamic deflection relationship for a simply supported (S–S) perforated nanobeam with varying filling ratio $\alpha$}
  \label{bookchfig3}
\end{figure}

The influence of filling ratio on bending behavior of perforated 
nanobeams with sinusoidal loads is illustrated in Fig. \ref{bookchfig3}. The main purpose is to capture the relationship between static and dynamic deflection from Fig. \ref{bookchfig3}. 
For a fixed filling ratio $\alpha$ (e.g., $\alpha = 0.2$), we can see that the relationship between dynamic and static responses exhibits a constant proportional behavior. Specifically, the ratio of dynamic response to static response remains constant for all $ X \in (0,1)$. 
Similarly, for different values of $\alpha$, different constant ratios are obtained.

\begin{figure}[H]
  \centering
  \begin{subfigure}[b]{0.49\linewidth}
    \includegraphics[width=\linewidth]{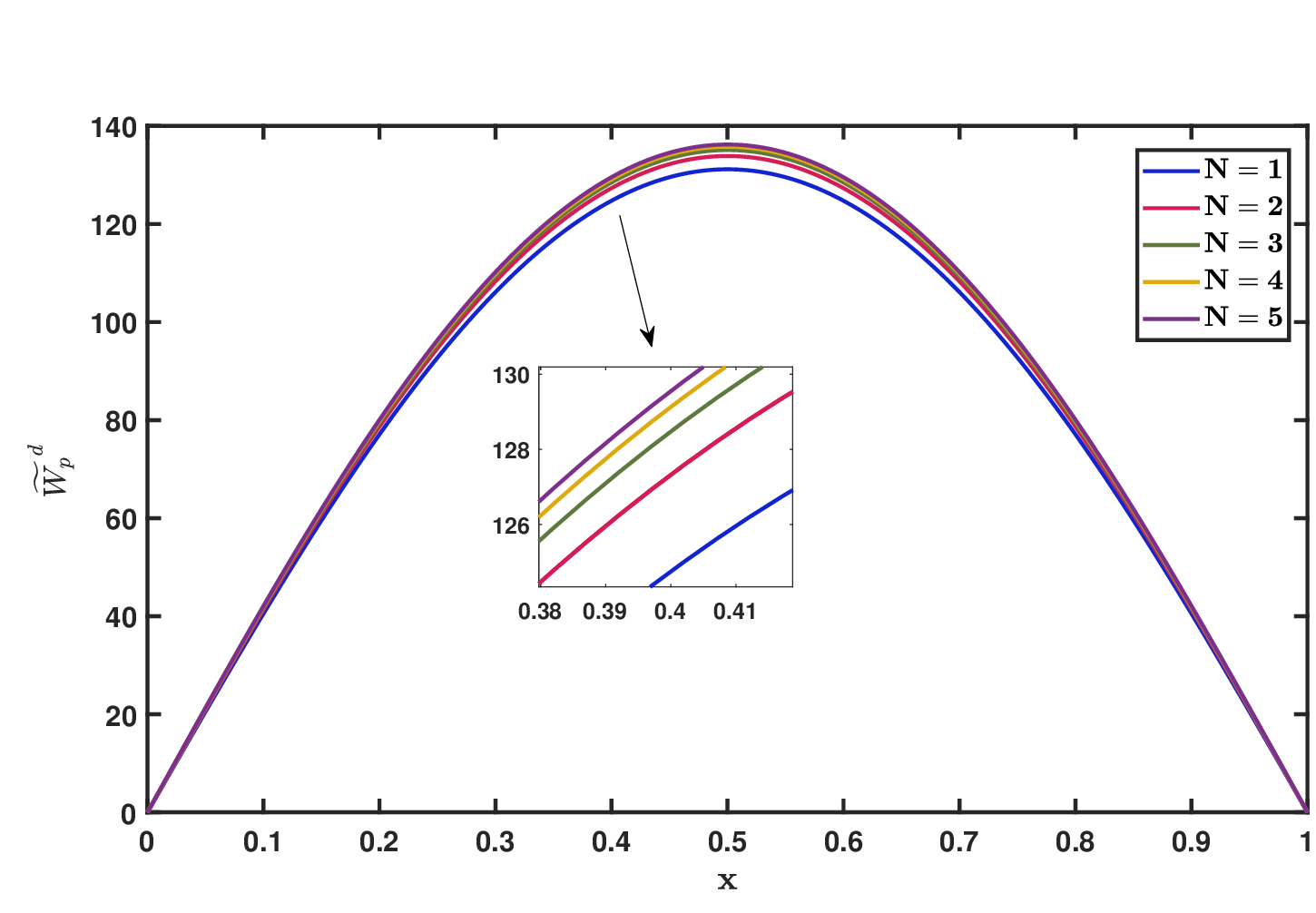}
     \caption{Variation of dynamic deflection with the number of perforations $N$}
  \end{subfigure}
  \begin{subfigure}[b]{0.49\linewidth}
    \includegraphics[width=\linewidth]{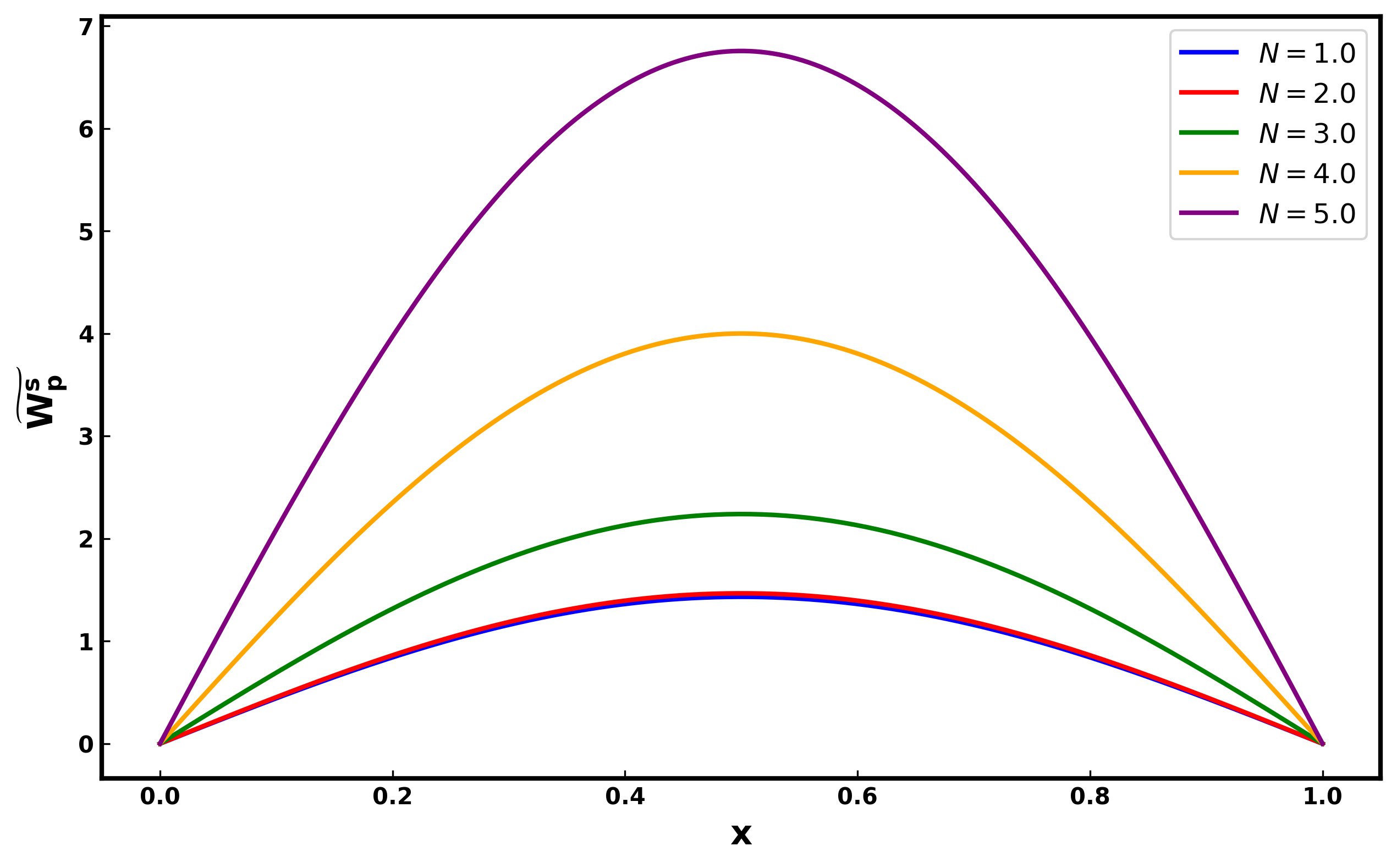}
    \caption{Variation of static deflection with the number of perforations $N$}
  \end{subfigure}
  \caption{Comparison of static and dynamic deflection for different numbers of holes $N$ in a simply supported perforated nanobeam with the parameter $\alpha = 0.5$ and $\bar \alpha = 0.2$}
  \label{bookchfig4}
\end{figure}

The number of perforations $N$ has a noticeable influence on bending response of the nanobeam. Fig. \ref{bookchfig4} reveals that as $N$ increases, more material is removed from the nanobeam, which reduces structural stiffness. Due to this reduction in stiffness, bending deflection increases in both static and dynamic conditions. 
Moreover, for a fixed value of $N$, the ratio between dynamic and static deflections remains constant for all $X \in (0,1)$ in Fig. \ref{bookchfig4}. When the number of holes changes, this ratio also changes, giving a different constant relation between static and dynamic responses for each value of $N$.

\begin{figure}[H]
  \centering
  \begin{subfigure}[b]{0.49\linewidth}
    \includegraphics[width=\linewidth]{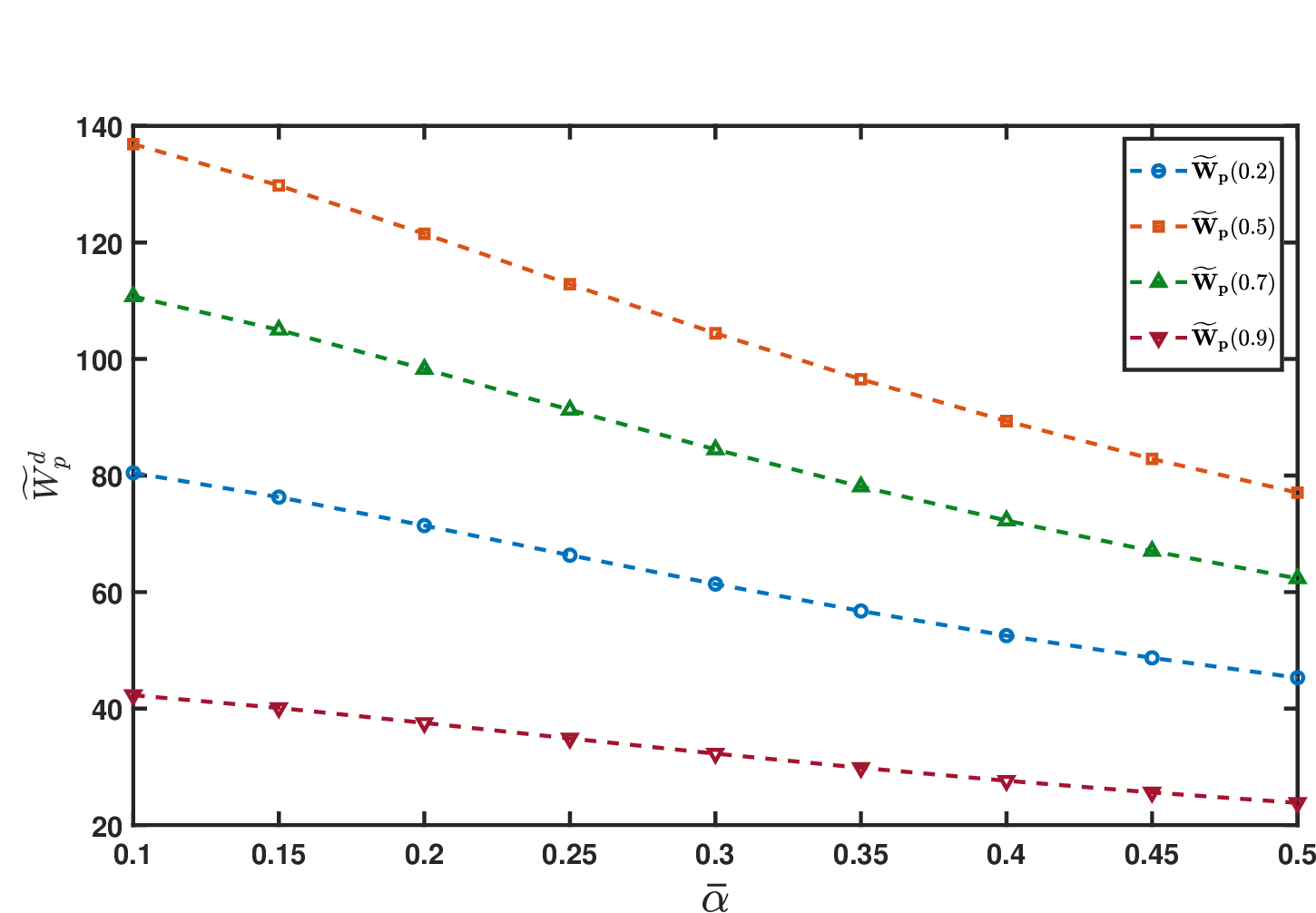}
     \caption{Variation of dynamic deflection with non-local parameter $\bar \alpha$}
  \end{subfigure}
  \begin{subfigure}[b]{0.49\linewidth}
    \includegraphics[width=\linewidth]{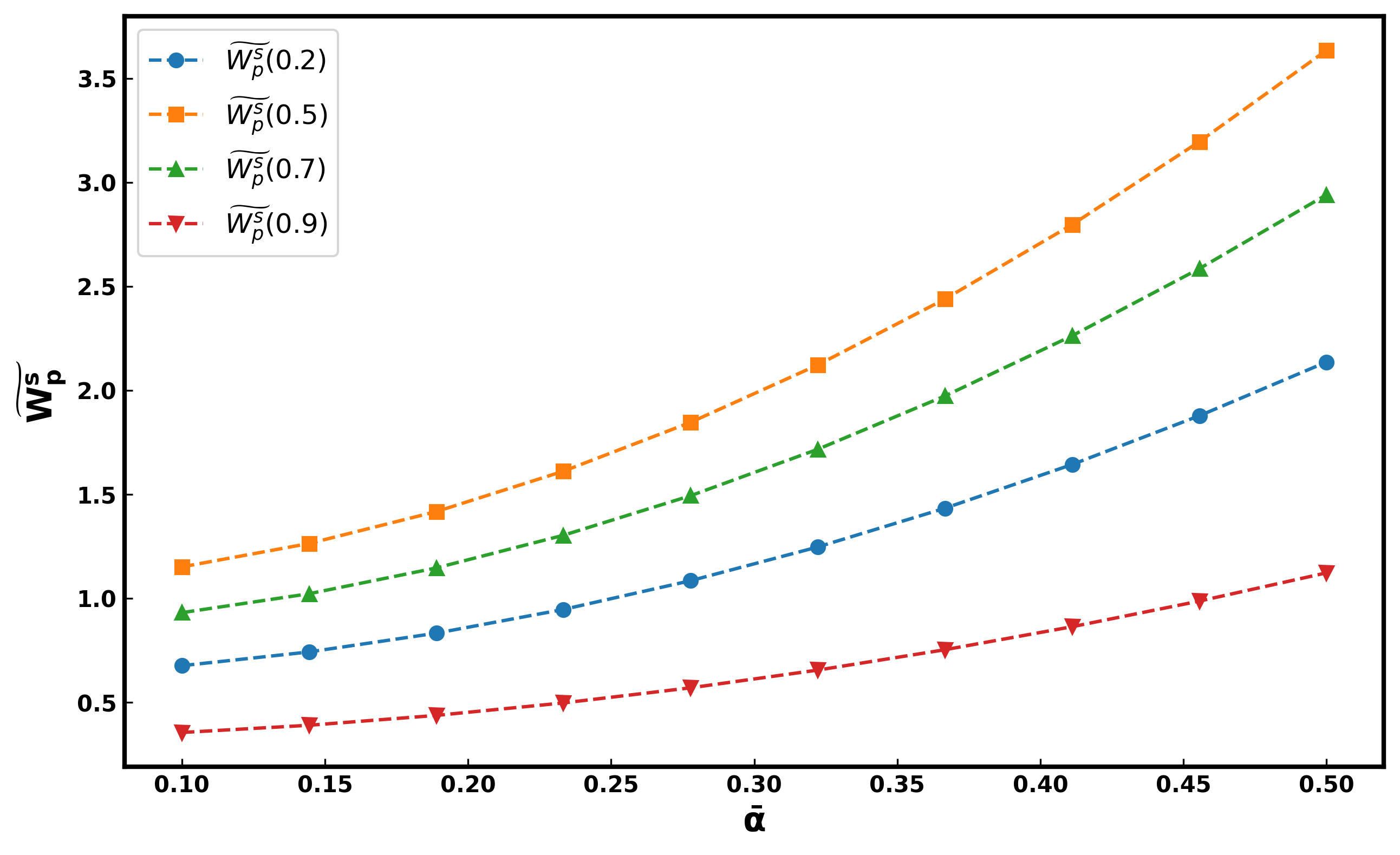}
    \caption{Variation of static deflection with non-local parameter $\bar \alpha$}
  \end{subfigure}
  \caption{Relationship between static and dynamic deflection of an S–S perforated nanobeam with variation in non-local parameter $\bar \alpha$ for the parameters $\alpha = 0.8 $ and $N=2$}
  \label{bookchfig5}
\end{figure}

It is worth mentioning that the non-local parameter $\bar{\alpha}$ has a very significant impact on static and dynamic deflection. In the static case, Fig. \ref{bookchfig5} illustrates that, as non-local parameter $\bar{\alpha}$ increases, the effective stiffness of the nanobeam decreases, which leads to an increase in static deflection. In contrast, for the dynamic case, an increase in $\bar{\alpha}$ decrease in dynamic deflection.

It also may be written that, for a fixed value of $\bar{\alpha}$, the ratio between dynamic and static deflections remains constant. In other words, the quantity
$\frac{\widetilde W_p^{\text{dynamic}}}{\widetilde W_p^{\text{static}}}$
is constant for a given $\bar{\alpha}$ for all $X \in (0,1)$. However, for different values of $\bar{\alpha}$, this constant takes different values.

\section{Conclusion}\label{sec:conclusion}
In this chapter, we analyze non-dimensional static and dynamic deflection and examine the relationship between them. The static deflection is obtained using the FL-TFC with Domain mapped method, which is computationally efficient, requires less computational time, and provides faster convergence compared to traditional ANN and PINN. In contrast, the dynamic deflection is obtained using the Galerkin method. 
Here, our focus has been to note the constant ratio between dynamic deflection and static deflection by taking different parameter values like filling ratio $\alpha$, number of rows of holes $(N)$, and nonlocal parameter $\bar{\alpha}$. For a given filling ratio $\alpha$, the relationship between dynamic and static responses exhibits a constant proportional behavior. Specifically, the ratio of dynamic response to static response remains constant in the total domain. Also, for a fixed number of holes $N$, the ratio between dynamic and static deflections remains constant across the domain but changes with different values of $N$, giving a distinct constant relationship for each case. Similarly, for a fixed nonlocal parameter $\bar{\alpha}$, the dynamic-to-static deflection ratio stays constant over the domain, while varying $\bar{\alpha}$ leads to different constant values.

It may be noted that when $N$ increases, more material is removed from the nanobeam, which reduces structural stiffness. Due to this reduction in stiffness, bending deflection increases under both static and dynamic conditions. 
The nonlocal parameter $\bar{\alpha}$ also has a significant effect on both static and dynamic deflections. In the static case, as the nonlocal parameter $\bar{\alpha}$ increases, the effective stiffness of the nanobeam decreases, which leads to an increase in static deflection. In contrast, for the dynamic case, an increase in $\bar{\alpha}$ leads to a decrease in dynamic deflection.

\subsection*{Acknowledgment}
The authors sincerely thank the National Institute of Technology Rourkela for providing the necessary research facilities. The first author also gratefully acknowledges the financial support received through a fellowship from CSIR, India.

%\section*{Statements and Declarations}
%No funding was received to assist with the preparation of this manuscript.
%\begin{itemize} 
%\item \textbf{Data:} Data sharing not applicable to this article as no datasets were generated or analysed during the current study.
%\item \textbf{Funding:} No funding was received to assist with the preparation of this manuscript.
%\item \textbf{Competing interests:} The authors have no competing interests to declare that are relevant to the content of this article.
%\end{itemize} 

%\section*{Acknowledgement}

%\bibliographystyle{plain}
\bibliography{reff}

@article{raissi2019physics,
  title={Physics-informed neural networks: A deep learning framework for solving forward and inverse problems involving nonlinear partial differential equations},
  author={Raissi, Maziar and Perdikaris, Paris and Karniadakis, George E},
  journal={Journal of Computational physics},
  volume={378},
  pages={686--707},
  year={2019},
  publisher={Elsevier}
}

@article{mortari2019multivariate,
  title={The multivariate theory of connections},
  author={Mortari, Daniele and Leake, Carl},
  journal={Mathematics},
  volume={7},
  number={3},
  pages={296},
  year={2019},
  publisher={MDPI AG}
}

@article{mortari2017theory,
  title={The theory of connections: Connecting points},
  author={Mortari, Daniele},
  journal={Mathematics},
  volume={5},
  number={4},
  pages={57},
  year={2017},
  publisher={MDPI}
}

@article{leake2020multivariate,
  title={The multivariate theory of functional connections: Theory, proofs, and application in partial differential equations},
  author={Leake, Carl and Johnston, Hunter and Mortari, Daniele},
  journal={Mathematics},
  volume={8},
  number={8},
  pages={1303},
  year={2020},
  publisher={MDPI}
}

@article{sahu2026physics,
  title={Physics-informed functional link with theory of functional connections technique for solving differential equations},
  author={Sahu, Iswari and Kumar, Sandeep and Chakraverty, S},
  journal={Neurocomputing},
  pages={132795},
  year={2026},
  publisher={Elsevier}
}

@article{mall2017single,
  title={Single layer Chebyshev neural network model for solving elliptic partial differential equations},
  author={Mall, Susmita and Chakraverty, Snehashish},
  journal={Neural Processing Letters},
  volume={45},
  pages={825--840},
  year={2017},
  publisher={Springer}
}

@article{paszke2017automatic,
  title={Automatic differentiation in pytorch},
  author={Paszke, Adam and Gross, Sam and Chintala, Soumith and Chanan, Gregory and Yang, Edward and DeVito, Zachary and Lin, Zeming and Desmaison, Alban and Antiga, Luca and Lerer, Adam},
  journal = {NIPS 2017 Autodiff Workshop},
  year={2017}
  }

@article{dan2025physics,
  title={Physics-Informed Neural Networks for Solving Free Vibration Response of Cables Considering Bending Stiffness},
  author={Dan, Danhui and Liao, Xia and Ge, Liangfu and Han, Fei},
  journal={Journal of Vibration Engineering \& Technologies},
  volume={13},
  number={8},
  pages={1--21},
  year={2025},
  publisher={Springer}
}

@article{li2024physics,
  title={Physics-informed neural networks for friction-involved nonsmooth dynamics problems},
  author={Li, Zilin and Bai, Jinshuai and Ouyang, Huajiang and Martelli, Saulo and Tang, Ming and Yang, Yang and Wei, Hongtao and Liu, Pan and Wei, Ronghan and Gu, Yuantong},
  journal={Nonlinear Dynamics},
  volume={112},
  number={9},
  pages={7159--7183},
  year={2024},
  publisher={Springer}
}

@article{martinez2026physics,
  title={Physics-informed neural networks with dynamical boundary constraints},
  author={Mart{\'\i}nez-Esteban, Andr{\'e}s and Calvo-Barl{\'e}s, Pablo and Mart{\'\i}n-Moreno, Luis and Rodrigo, Sergio G},
  journal={Communications in Nonlinear Science and Numerical Simulation},
  pages={109866},
  year={2026},
  publisher={Elsevier}
}

@article{leake2020deep,
  title={Deep theory of functional connections: A new method for estimating the solutions of partial differential equations},
  author={Leake, Carl and Mortari, Daniele},
  journal={Machine learning and knowledge extraction},
  volume={2},
  number={1},
  pages={37--55},
  year={2020},
  publisher={MDPI}
}

@article{schiassi2020extreme,
  title={Extreme theory of functional connections: A physics-informed neural network method for solving parametric differential equations},
  author={Schiassi, Enrico and Leake, Carl and De Florio, Mario and Johnston, Hunter and Furfaro, Roberto and Mortari, Daniele},
  journal={arXiv preprint arXiv:2005.10632},
  year={2020}
}

@article{luschi2012simple,
	title={A simple analytical model for the resonance frequency of perforated beams},
	author={Luschi, Luca and Pieri, Francesco},
	journal={Procedia Engineering},
	volume={47},
	pages={1093--1096},
	year={2012},
	publisher={Elsevier}
	
}

@article{luschi2014analytical,
	title={An analytical model for the determination of resonance frequencies of perforated beams},
	author={Luschi, Luca and Pieri, Francesco},
	journal={Journal of Micromechanics and Microengineering},
	volume={24},
	number={5},
	pages={055004},
	year={2014},
	publisher={IOP Publishing}
}

@article{vadivelu2025flexural,
  title={Flexural behavior of perforated rectangular hollow section cold-formed steel beams: An experimental and numerical study},
  author={Vadivelu, Chitrarasu and Prabaharan, Vaishnavi and Akhas, Punitha Kumar},
  journal={Results in Engineering},
  volume={26},
  pages={104821},
  year={2025},
  publisher={Elsevier}
}

@article{abdelrahman2022static,
  title={Static bending of perforated nanobeams including surface energy and microstructure effects},
  author={Abdelrahman, Alaa A and Mohamed, Norhan A and Eltaher, Mohamed A},
  journal={Engineering with Computers},
  volume={38},
  number={Suppl 1},
  pages={415--435},
  year={2022},
  publisher={Springer}
}

@article{fallah2024physics,
  title={Physics-informed neural network for bending and free vibration analysis of three-dimensional functionally graded porous beam resting on elastic foundation},
  author={Fallah, Ali and Aghdam, Mohammad Mohammadi},
  journal={Engineering with Computers},
  volume={40},
  number={1},
  pages={437--454},
  year={2024},
  publisher={Springer}
}

@article{reddy2007nonlocal,
  title={Nonlocal theories for bending, buckling and vibration of beams},
  author={Reddy, JN1213},
  journal={International journal of engineering science},
  volume={45},
  number={2-8},
  pages={288--307},
  year={2007},
  publisher={Elsevier}
}

@article{fernandez2017vibrations,
  title={Vibrations of Bernoulli-Euler beams using the two-phase nonlocal elasticity theory},
  author={Fern{\'a}ndez-S{\'a}ez, Jos{\'e} and Zaera, R36797671423},
  journal={International Journal of Engineering Science},
  volume={119},
  pages={232--248},
  year={2017},
  publisher={Elsevier}
}

@article{kafkas2025size,
  title={Size-Dependent Bending Behavior of Perforated Nanobeams on Winkler-Pasternak Foundation.},
  author={Kafkas, U{\u{g}}ur},
  journal={International Journal of Engineering \& Applied Sciences (1309-0267)},
  volume={17},
  number={1},
  year={2025}
}

@article{garai2026effect,
  title={Effect of Axial Functional Gradation and Periodic Square Perforations on the Vibration of Nanobeams with Elastic Foundation},
  author={Garai, Ramanath and Gartia, Akash Kumar and Chakraverty, S},
  journal={International Journal of Structural Stability and Dynamics},
  pages={2750188},
  year={2026},
  publisher={World Scientific}
}

@article{eltaher2018resonance,
  title={Resonance frequencies of size dependent perforated nonlocal nanobeam},
  author={Eltaher, MA and Abdraboh, AM and Almitani, KH},
  journal={Microsystem Technologies},
  volume={24},
  number={9},
  pages={3925--3937},
  year={2018},
  publisher={Springer}
}

@INPROCEEDINGS{10389876,
  author={Kumar, Sandeep and Sahoo, Arup Kumar and Chakraverty, S.},
  booktitle={2023 International Conference on Ambient Intelligence, Knowledge Informatics and Industrial Electronics (AIKIIE)}, 
  title={Physics-informed Machine Learning Framework for Approximating the Modified Degasperis-Procesi Equation}, 
  year={2023},
  volume={},
  number={},
  pages={1-6}
}

@article{sahoo2025physics,
  title={Physics-informed neural network for vibration analysis of large membranes},
  author={Sahoo, Arup Kumar and Kumar, Sandeep and Chakraverty, S},
  journal={Journal of Nonlinear, Complex and Data Science},
  volume={25},
  number={7-8},
  pages={505--521},
  year={2025},
  publisher={De Gruyter}
}

@article{mortari2019high,
  title={High accuracy least-squares solutions of nonlinear differential equations},
  author={Mortari, Daniele and Johnston, Hunter and Smith, Lidia},
  journal={Journal of computational and applied mathematics},
  volume={352},
  pages={293--307},
  year={2019},
  publisher={Elsevier}
  }

@article{sahu2026physics1,
  title={Physics-Informed Functional Link Constrained Framework with Domain Mapping for Solving Bending Analysis of an Exponentially Loaded Perforated Beam},
  author={Sahu, Iswari and Garai, Ramanath and Chakraverty, S},
  journal={arXiv preprint arXiv:2604.07025},
  year={2026}
}

@article{Eringen1983,
author = {A. Cernal Eringen},
title = {On differential equations of nonlocal elasticity and solutions of screw dislocation and surface waves },
journal = {Journal of Applied Physics},
volume = {54},
number = {9},
pages = {4703–4710},
year = {1983}
}

\end{document}